\documentclass{article}
\usepackage{spconf,amsmath,epsfig}
\usepackage{amsfonts}
\usepackage{xcolor}

\let\OLDthebibliography\thebibliography
\renewcommand\thebibliography[1]{
  \OLDthebibliography{#1}
  \setlength{\parskip}{0pt}
  \setlength{\itemsep}{0pt plus 0.3ex}
}

\begin{document}\sloppy

\def\x{{\mathbf x}}
\def\L{{\cal L}}

\title{Package Theft Detection from Smart Home Security Cameras}
%
\name{Hung-Min Hsu$^{\ast}$, Xinyu Yuan$^{\ast}$, Baohua Zhu$^{\dagger,\ast}$, Zhongwei Cheng$^{\ast}$ and Lin Chen$^{\ast}$}
\address{$^{\ast}$WYZE Labs AI Team, Kirkland, WA, USA; $^{\dagger}$University of Washington, Seattle, WA, USA \\ \{hhsu,xyuan,zchen,lchen\}@wyze.com; bz22@uw.edu}

\maketitle

\begin{abstract}
Package theft detection has been a challenging task mainly due to lack of training data and a wide variety of package theft cases in reality. In this paper, we propose a new Global and Local Fusion Package Theft Detection Embedding (GLF-PTDE) framework to generate package theft scores for each segment within a video to fulfill the real-world requirements on package theft detection. Moreover, we construct a novel Package Theft Detection dataset to facilitate the research on this task. Our method achieves 80\% AUC performance on the newly proposed dataset, showing the effectiveness of the proposed GLF-PTDE framework and its robustness in different real scenes for package theft detection.
\end{abstract}

\section{Introduction}
\label{sec:intro}

During the COVID-19 pandemic, the significant increase of online shopping drives the rapid growth of e-commerce. According to the 2020 package theft statistics report\footnote{https://www.crresearch.com/blog/2020-package-theft-statistics-report}, the package stolen rate has increased from 36\% in 2019 to 43\% in 2020. Especially, 64\% of respondents have been stolen more than once. 
Approximately 144 million customers were affected with an average loss per household of 106 USD. The motivation of this work is to build an effective and practical solution to detect the package theft events, leveraging widely adopted security cameras.

Developing the intelligent computer vision system of automatic package theft detection (PTD) is necessary to alleviate the labor and time of the society. Inspired by anomaly detection system, the target of PTD can be considered to extract the patterns in a specific time period window of the package stealing behavior. In another word, PTD is a special case of video understanding that can be handled by differentiating package stealing behavior from normal patterns.


Real-world package theft events are complicated since the environment and human behavior are diverse and varied. It is impractical to exhaust all possible package theft event patterns. Therefore, it is essential to detect the abnormal package pickup behavior rather than modeling prior information on theft behaviors.

Training a neural network for PTD is challenging in practice. It's not reliable to simply use the labeled normal patterns in the training data, and then expect any other patterns in the real-world videos deviating from the trained model as the package theft event. Specifically, it is impossible to pre-define all possible package pickup/delivery behaviors. Moreover, the boundary between a normal and an anomalous behavior for package pickup/delivery is ambiguous, therefore it is debatable to purely use the training data to detect package theft. In this paper, we propose a novel package theft detection framework by using the weakly labeled training videos and human pose information. Our contributions in this work are summarized as follows,
\begin{itemize}
  \item Propose PTD system with new Global and Local Fusion Package Theft Detection Embedding framework (GLF-PTDE).
  \item Introduce the human pose information into PTD solution to improve performance.
  \item Build a novel package theft detection dataset to facilitate PTD research.
\end{itemize}
Based on the experimental results, the proposed neural network can identify the package theft segments in a video based on the high anomaly scores. 


\section{Related Works}\label{sec:related_works}

The most related work of PTD is anomaly detection in the computer vision community including human behavior and traffic monitoring \cite{wang2019anomaly}. Nowadays, due to the success of deep learning technologies, a larger amount of neural networks are proposed to deal with various application tasks. For example, \cite{sabokrou2017deep} applies 3D deep neural networks for crowded scene object detection and localization. 
The most popular anomaly detection scenario is human violence or abnormal event detection crowd scene. 
\cite{sultani2018real} proposes a neural network to generate the anomaly score for videos. Then, \cite{xu2017detecting} uses a double fusion framework to integrate the appearance and motion features for anomaly detection. 



\section{The Proposed Method}\label{sec:method}

\begin{figure*}[t]
\begin{center}
   \includegraphics[width=0.65\linewidth]{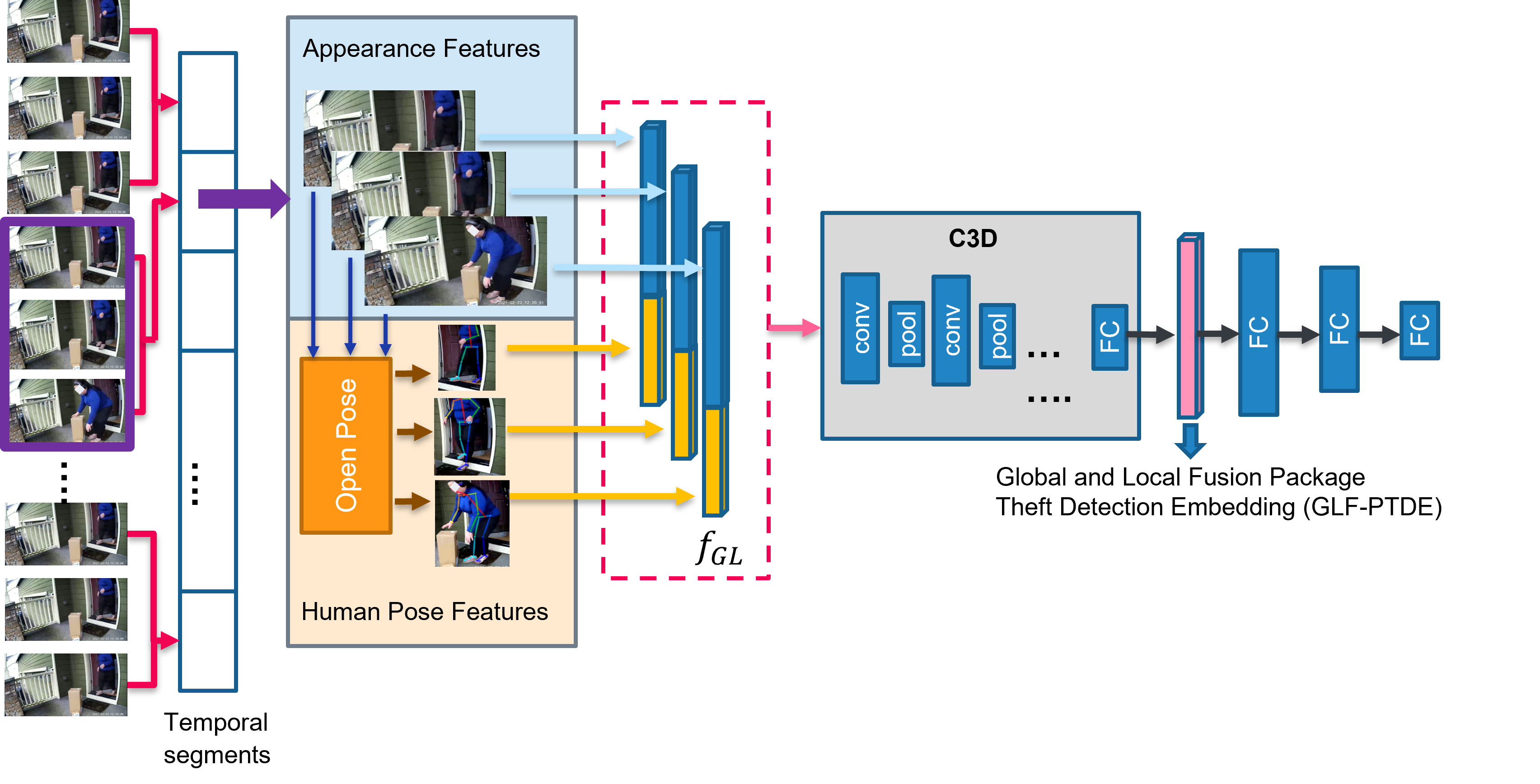}
\end{center}
   \caption{The framework of proposed Package Theft Detection solution. Firstly, the input video is split into small segments, and the appearance features and human pose features are generated for each frame. After that, the concatenation operation is applied to combine these two features as the input of C3D to generate Global and Local Fusion Package Theft Detection Embedding (GLF-PTDE). Finally, three FC layers are applied to generate the package theft confidence score for each segment.}
\label{fig:framework}
\end{figure*}


\subsection{Global Package Theft Detection Embedding}

For global feature extraction, we use the whole image sequences as the inputs called as Global Package Theft Detection Embedding (G-PTDE). G-PTDE considers the foreground and background information to determine the segments are package stealing or not. Thus, we can express G-PTDE ${f}_{G-PTDE}$ as:
\begin{equation}
{f}_{G-PTDE}(X) = 
 \mathbf{\phi}(X) \in \mathbb{R}^{1 \times D},
\end{equation}
where $\mathbf{\phi}$ denotes C3D convolution operation \cite{tran2015learning}, which is applied to aggregate the temporal information into the segment-level embedding. $D$ is the dimension of C3D output. The video is divided into several segments, and then $X$ is the set of appearance features of each image in a segment. Therefore, ${f}_{G-PTDE}$ is segment-level embedding, for each segment we generate the corresponding PTD score.

\subsection{Local Package Theft Detection Embedding}

In our framework, we not only consider the appearance embedding feature but also the human pose features since the human pose can be used to detect package theft semantically. We use OpenPose \cite{cao2019openpose} to obtain the human pose information, which is trained on COCO dataset to generate the human pose features. These inferred human pose 2D keypoints (body joints) are referred to as the Local Package Theft Detection Embedding (L-PTDE). Unlike the G-PTDE also considers the background information, L-PTDE is the representation of foreground information so that it only pays attention to the local information. L-PTDE feature consists of 18 joints of each identity in each frame, whose dimension is $18 \times 3$. The $3$ comes from the (x, y) 2D joint coordinates and the confidence score. 
Following the same processing as $f_{G-PTDE}$, the video is divided into several segments.
Assume the segment size is $L$, $S = \left \lfloor \frac{T}{L} \right \rfloor$ is the number of segments, and $D^ \prime$ is the dimension of the segment-level feature. $T$ is the length of the video. Therefore, the L-PTDE ${f}_{K}^{segment}$ can be defined as the follows:
\begin{equation}
{f}_{K}^{segment} = \{f_{K}^{1}, \cdots , f_{K}^{S}\} \in \mathbb{R}^{S \times L \times D^ \prime}.
\end{equation}




\subsection{Global and Local Fusion Package Theft Detection Embedding}

Most anomaly detection systems only focus on the entire image, which is easily influenced by the background information. Therefore, the obvious flaw is that the information of the human pose is missing. The package theft detection should pay more attention to the walking posture to enhance the foreground information. In our system, we incorporate the global and local features to generate a robust embedding $f_{GLF-PTDE}$ by concatenating these two features as the input for the C3D network so that we involve both the original entire image features from the whole video sequence, and features of different human joints, e.g., head, shoulder, hip, legs. Fig.\ref{fig:framework} shows the entire pipeline of GLF-PTDE.

The proposed Global and Local Fusion Package Theft Detection Embedding (GLF-PTDE) $f_{GLF-PTDE}$ is defined as the follows: 
\begin{equation}
\begin{split}
{f}_{GLF-PTDE} &=  \mathbf{\phi_{GLF}}(X^i \oplus f_{K}^{segment,i}) \in \mathbb{R}^{S \times D^{\prime \prime}},
\end{split}
\end{equation}
where ${f}_{K}^{segment,i}$ denotes the human pose feature; $\mathbf{\phi_{GLF}}$ is a C3D convolution operation for global and local information fusion; $\oplus$ represents the concatenation operation; $i$ denotes the index of segment; $S$ is the number of segments; $D^{\prime \prime}$ means the dimension of GLF-PTDE.

\subsection{Loss function}

The same as anomaly detection, the definition of package theft detection is difficult and subjective. On the other hand, the representative of training data is not sufficient to represent all the package theft patterns. Thus, our package detection is defined as low likelihood pattern detection instead of binary classification problem. Inspired by \cite{sultani2018real}, we also treat the PTD as a regression problem, which means that we aim to determine the segments of package stealing to have the highest anomaly scores. For the regression problem, we have to generate the embedding for the regression head. Since we aim to generate the highest scores for package stealing
video segments, it is straightforward to apply the ranking loss to train the embedding. 

\begin{equation}
f(S_{pt}) > f(S_n)
\end{equation}
where $S_{pt}$ and $S_n$ represent package theft and normal video
segments, $f(S_{pt})$ and $f(S_n)$ represent the corresponding
predicted scores, respectively. The ranking function should be modified to satisfy the segment-level annotations, therefore, the rank function is defined as follows:

\begin{equation}
\max _{i \in B_a}f(S_{pt}^i) > \max _{i \in B_a}f(S_n^i)
\end{equation}
where $B_a$ means the set of instances in the dataset.

The loss function is followed the time structure of package theft videos since the real-world package stealing behaviors occur in a short period of time so that the package theft scores are sparse. Moreover, the input video is split into many segments, the package theft scores in consecutive segments should be smooth. Therefore similar to \cite{sultani2018real}, we minimize the difference of scores for adjacent video segments, and then the temporal smoothness can be maintained. The definition of loss function is as follows:

\begin{equation}
\begin{split}
\mathcal{L}= max(0, 1 -\max _{i \in B_a}f(S_{pt}^i) + \max _{i \in B^i_a}f(S_n^i)) \\
+ \lambda_1 \sum ^{(n-1)}_i (f(S_{pt}^i) - f(S_{pt}^{i+1}))^2 + \lambda_2 \sum ^{n}_{i} f(S_{pt}^i)
\end{split}
\end{equation}
where $max(0, 1 -\max _{i \in B_a}f(S_{pt}^i) + \max _{i \in B_a}f(S_n^i))$ is the ranking loss for multiple instance learning; $\sum ^{n}_{i} f(S_{pt}^i)$ means the sparsity  and $\sum ^{(n-1)}_i (f(S_{pt}^i) - f(S_{pt}^{i+1}))^2 $ represents the temporal smoothness term. 


\section{Experiments}\label{sec:results}

\subsection{Dataset}

There is no existing dataset specific for package theft detection. We construct a new one for evaluation, which consists of 120 surveillance videos with an average video length of about 52 seconds (28.8 frames per second) and covers four categories: package theft, normal pickup, normal delivery and irrelevant. The category distribution on video numbers is shown in Table~\ref{tab:data}. All videos are annotated manually by human.

\begin{table}[h]
\footnotesize
\begin{center}
\resizebox{85mm}{7mm}{
\begin{tabular}{c|c|c c c}
\hline 
 & Package Theft & Pickup & Delivery & Irrelevant  \\
\hline 
 training & 60 & 20  & 20 & 20  \\
\hline
 testing & 40 & 10  & 20 & 10  \\
\hline 
\end{tabular}
}
\end{center}
\caption{The distribution of video categories in package theft dataset. There are 120 videos for training and 80 for testing, respectively.}
\label{tab:data}
\end{table}

\subsection{Implementation Details}

\begin{figure}[h]
\begin{center}
  \includegraphics[width=0.65\linewidth]{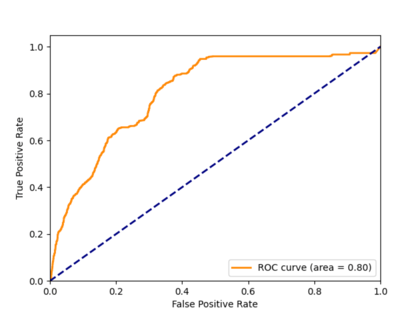}
\end{center}
  \caption{Receiver Operating Characteristic curve (ROC) of GLF-PTDE.}
\label{fig:roc_glf}
\end{figure}

We extract the GLF-PTDE to detect the package theft, which is extracted from the fully connected (FC)
layer FC6 of the C3D network \cite{tran2015learning}. All of the videos are re-sized to $240 \times 320$ and adjusted to 30 frames per second. For each segment, we use C3D to calculate the 16-frame video clip feature followed by L2 normalization, and then we take the average of all 16-frame clip features to generate the segment feature. We concatenate the CNN feature and human pose feature as the input of C3D, and the human pose feature is the normalized image coordinate of each human joint and the corresponding confidence score. Here we use $18$ joints so the human pose feature size is $18 \times 3$. The pose features are extracted by the COCO dataset pre-trained model since we do not have the human pose ground truth in our package theft dataset. This pseudo pose feature is sufficient enough to improve the overall performance, however, fine-tuning on the pose model should be useful to obtain further improvement. After C3D feature extraction, we obtain a 4096-D feature vector and it is past to a 3-layer FC neural network to estimate the package theft confidence score. The three FC layers are 512, 32 and 1, respectively. The training process is conducted as training from scratch. The optimizer is Adagrad with learning rate 0.01 and the training epochs is 5000. The Receiver Operating Characteristic curve (ROC) of G-PTDE is shown in the Fig~\ref{fig:roc_glf}, based on which the threshold of package theft detection sets at $0.2$. The parameters of sparsity and smoothness constraints are both $8 \times 10^{-5}$. There is no overlapping frames between segments for efficiency consideration.

\subsection{Evaluation and Results}

In our experiments, we compare our method with the state-of-the-art approach of anomaly detection \cite{sultani2018real}, which is noted as G-PTDE in this paper. In terms of metric, we use the corresponding area under the curve (AUC) to evaluate the performance of our method since only a small portion of a long video contains package theft. We conduct three experiments settings in Table~\ref{tab:res}. First, we use G-PTDE for package theft detection by the UCF Crime dataset pre-trained model, which is trained by 400 videos including multiple different types of crime categories such as abuse, arrest, arson, assault, burglary, explosion, fighting, road accident, robbery, shooting, shoplifting, and stealing. The AUC of G-PTDE with the UCF-Crime pre-trained model is 68\%. If the G-PTDE is trained on our package theft dataset from scratch, the AUC can achieve 76\%. As for our GLF-PTDE, the AUC can achieve 80\%, which reveals that considering the human pose information is effective to improve performance with the increase of AUC by 4\% (also trained from scratch), and it shows the importance of adding the foreground information.
	
\begin{table}[h]
\footnotesize
\begin{center}
\vspace{0.3em}
\resizebox{85mm}{10mm}{
\begin{tabular}{c|c|c c c}
\hline 
 Training Data & video \# & AUC  \\
\hline 
UCF Crime dataset (G-PTDE \cite{sultani2018real}) & 400 & 0.68  \\
\hline 
 Package Theft dataset (G-PTDE \cite{sultani2018real}) & 120 & 0.76 \\
 \hline
 Package Theft dataset (GLF-PTDE) & 120 & 0.80 \\
 \hline
\end{tabular}
}
\end{center}
\caption{PTD performance on different training datasets}
\label{tab:res}
\end{table}

Moreover, we also dive into the performance analysis in three different normal condition comparisons: normal delivery, normal pickup, and irrelevant. According to Table~\ref{tab:res_normal}, it shows that the significance of performance improvement of GLF-PTDE is from the normal delivery versus package theft. GLF-PTDE is obviously having more capability of distinguishing for the normal delivery and package theft, which is important for package theft detection system. The improvement brought by GLF-PTDE is 14\% on normal delivery, and 1\% and 2\% for the normal pickup and irrelevant actions, respectively.
\begin{table}[h]
\footnotesize
\begin{center}
\vspace{0.3em}
\resizebox{85mm}{7mm}{
\begin{tabular}{c|c|c c c}
\hline 
 & Method & Delivery & Pickup & Irrelevant  \\
\hline 
 Package Theft & G-PTDE \cite{sultani2018real} & 0.54  & 0.67 & 0.85  \\
 & GLF-PTDE & 0.68  & 0.68 & 0.87 \\
 \hline
\end{tabular}
}
\end{center}
\caption{PTD performance on different normal conditions}
\label{tab:res_normal}
\end{table}


\section{Conclusion}\label{sec:conclusion}

We propose a deep learning approach to detect package theft for home surveillance videos. Due to the complexity of these realistic package theft behaviors, we propose Global and Local Fusion Package Theft Detection Embedding (GLF-PTDE) framework for package theft detection. 
Moreover, we propose a novel package theft benchmark, which is new large-scale package theft detection dataset including a variety of real-world package theft scenarios. The experimental results on the proposed package theft detection dataset show that
our proposed package theft detection approach performs significantly better than baseline methods. 

\bibliographystyle{IEEEbib}
\bibliography{icme2022template}

\end{document}